\documentclass[journal,draftcls,onecolumn]{IEEEtran}

\usepackage{graphics}
\usepackage{graphicx}
\usepackage[linesnumbered,ruled,vlined]{algorithm2e}
\usepackage{algorithmicx}
\usepackage{algpseudocode}
\usepackage[]{algorithm2e}
\usepackage{amsfonts}
\usepackage{amssymb}
\usepackage{amsmath}
\usepackage{lipsum}
\usepackage{bm}
\usepackage{bbm}
\usepackage{subfig}
\usepackage{array}
\usepackage{breqn}
\usepackage{epstopdf}
\usepackage{booktabs}
\usepackage{siunitx}

\usepackage{lineno}
\usepackage{adjustbox}
\usepackage{multirow}
\usepackage{color}
\usepackage[noabbrev]{cleveref}
\usepackage{cite}

\usepackage[normalem]{ulem}

\RequirePackage{snapshot}

\graphicspath{{./Figures/}}

\begin{document}
%
\title{Unsupervised Classification of PolSAR Data Using a Scattering Similarity Measure Derived from a Geodesic Distance}
%
%
%

\author{Debanshu~Ratha~\IEEEmembership{Student~Member,~IEEE}, Avik~Bhattacharya,~\IEEEmembership{Senior~Member,~IEEE} \\ and Alejandro~C.~Frery,~\IEEEmembership{Senior Member,~IEEE}
\thanks{D.~Ratha and A.~Bhattacharya are with the Center of Studies in Resources Engineering, Indian Institute of Technology Bombay, Mumbai, India (\mbox{e-mail:~avikb@csre.iitb.ac.in}).}
\thanks{A.~C.~Frery is with the Universidade Federal de Alagoas, Macei\'o, Brazil}%
}

\maketitle

\begin{abstract}
In this letter, we propose a novel technique for obtaining scattering components from Polarimetric Synthetic Aperture Radar (PolSAR) data using the geodesic distance on the unit sphere. This geodesic distance is obtained between an elementary target and the observed Kennaugh matrix, and it is further utilized to compute a similarity measure between scattering mechanisms. The normalized similarity measure for each elementary target is then modulated with the total scattering power $(\mathrm{Span})$. This measure is used to categorize pixels into three categories i.e. odd-bounce, double-bounce and volume, depending on which of the above scattering mechanisms dominate. Then the maximum likelihood classifier of Lee et al.~\cite{Lee2004} based on the complex Wishart distribution is iteratively used for each category. Dominant scattering mechanisms are thus preserved in this classification scheme. We show results for L-band AIRSAR and ALOS-2 datasets acquired over San-Francisco and Mumbai, respectively. The scattering mechanisms are better preserved using the proposed methodology than the unsupervised classification results using the Freeman-Durden scattering powers on an orientation angle (OA) corrected PolSAR image. Furthermore, (1)~the scattering similarity is a completely non-negative quantity unlike the negative powers that might occur in double-bounce and odd-bounce scattering component under Freeman Durden decomposition (FDD), and (2)~the methodology can be extended to more canonical targets as well as for bistatic scattering.    
\end{abstract}


%
\IEEEpeerreviewmaketitle

\section{Introduction}
\IEEEPARstart{P}{olarimetric} SAR (PolSAR) image classification is an important tool for land use/land cover mapping. In this context several supervised and unsupervised techniques have been reported in the literature~\cite{kong1988identification,lee1994classification,ferro2001unsupervised,FreryCorreiaFreitas:ClassifMultifrequency:IEEE:2007,Formont2011,Horta2008,Fernnndez-Michelli,Doulgeris2015,vanzyl_89,Cloude97,pottier1997application,Cao2007,Wang2013,Chen2013}. Some of these techniques are based on statistical characteristics of PolSAR data while others are based on the physical scattering process. However, there has been growing interest in the classification of PolSAR images using a hybrid approach: statistical analysis combined with target scattering properties. 

Kong et al.~\cite{kong1988identification} first proposed a maximum-likelihood (ML) classification with a probabilistic distance measure based on the Gaussian distribution, while Lee et al.~\cite{lee1994classification} proposed ML classification based on the complex Wishart distribution. Ferro-Famil et al.~\cite{ferro2001unsupervised} extended it to multi-frequency PolSAR data, and Frery et al.~\cite{FreryCorreiaFreitas:ClassifMultifrequency:IEEE:2007} incorporated spatial evidence. Formont et al. proposed a classification procedure of PolSAR data in heterogeneous clutter based on the Spherically Invariant Random Vector (SIRV) model~\cite{Formont2011}. In~\cite{Horta2008,Fernnndez-Michelli} Horta et al. and Fern\'andez-Michelli et al. respectively proposed classification of PolSAR images using a mixture of $\mathcal{G}^{0}_{P}$ while Doulgeris proposed a classification based on the $\mathcal{U}_{d}$ distribution~\cite{Doulgeris2015}. 

van Zyl~\cite{vanzyl_89} proposed a simple unsupervised classification scheme which compares scattering mechanisms of a pixel with elementary scattering such as even-bounce, odd-bounce and diffused. Cloude and Pottier~\cite{Cloude97} used an eigenvalue-eigenvector decomposition of the coherency matrix to obtain the scattering entropy ($H$) and mean the scattering type ($\underline{\alpha}$), which are then used to segment a PolSAR image into eight clusters.  
Subsequently, in~\cite{pottier1997application} the segmentation was extended into sixteen clusters by adding the anisotropy $A$. Following this, several methods have been developed which utilize additional polarimetric parameters and similarity measures to improve the classification accuracy~\cite{Cao2007, Wang2013,Chen2013}. 

In the context of a hybrid approach for unsupervised PolSAR image classification, the method proposed by Lee et al.~\cite{Lee2004}, utilizing the Freeman-Durden scattering power decomposition (FDD)~\cite{freeman98} is widely used. In this method, a pixel is categorized into three scattering categories: odd-bounce, double-bounce and volume obtained from the FDD. Scattering purity is conserved while clustering pixels within each category, and finally an iterative Wishart classification scheme is applied. In subsequent studies, the FDD was replaced by other model-based scattering power decomposition methods. These methods try to circumvent the problem of overestimating the volume scattering power. However, most of these methods have high implementation and computational complexity~\cite{Chen_Advances_and_perspectives_2014}. 

In this letter, a novel technique is proposed to obtain scattering components from PolSAR data using a geodesic distance on the unit sphere~\cite{Ratha_CD_2017}. This geodesic distance is obtained between an elementary target and the observed Kennaugh matrix. It is further utilized to compute a similarity measure between scattering mechanisms. The normalized similarity measure for each elementary target is then modulated with the total scattering power $(\mathrm{Span})$ to obtain individual canonical scattering components. These scattering components are useful in labeling the pixels into three categories: odd-bounce, double-bounce and volume, depending on the dominant scattering mechanism. Then, the maximum likelihood classifier of Lee et al.~\cite{Lee2004} based on the complex Wishart distribution is iteratively used for each category. Dominant scattering mechanisms are, hence, preserved in this classification scheme. Several advantages of the proposed method to compute the scattering component similarity, are highlighted in this work. 

\section{Methodology}
\subsection{Geodesic Distance}
A radar target is characterized by a scattering (or Sinclair) matrix $\mathbf{S}$ which describes dependence of its scattering properties on the polarization. It is defined in the HV (H: Horizontal and V: Vertical polarization) basis as
\begin{equation}
\mathbf{S}=
\left[\begin{array}{cc}
S_{\text{HH}} & S_{\text{HV}}\\
S_{\text{VH}} & S_{\text{VV}}
\end{array}\right]
\label{Eq:scattering_matrix}
\end{equation}
where each element is a complex quantity: the amplitude and the phase of the scattered electromagnetic (EM) signal. For a monostatic radar, $\mathbf{S}$ is assumed to be symmetric, i.e. $S_{\text{HV}}=S_{\text{VH}}$. The Pauli vector is an equivalent form of representing the same information as the Sinclair matrix and is defined as $\mathbf{k} = \frac{1}{\sqrt{2}}[S_{\text{HH}} + S_{\text{VV}},\; S_{\text{HH}}-S_{\text{VV}},\; 2S_{\text{HV}}]^T$, where the superscript $T$ denotes transposition. Other special matrices in PolSAR theory are derived from $\mathbf{S}$. The coherency matrix $\mathbf{T}$ is an incoherent measurement obtained by the process of multi-looking:
\begin{equation}
\mathbf{T} = (T_{i,j})_{1\leq i,j\leq 3} = \frac{1}{L}\sum_{i=1}^{L}\mathbf{k}_i\mathbf{k}_i^{*T}
\end{equation}
where superscript $*$ denotes the complex conjugate, and $L$ is the number of looks. By definition, the coherency matrix is Hermitian. 

The $4 \times 4$ real symmetric Kennaugh matrix in monostatic configured PolSAR conveys the information about the transformation of incident and received Stokes vector. For the coherent case, the matrix $\mathbf{K}$ can be obtained from $\mathbf{S}$ in the following manner~\cite{You2017} :  
\begin{equation}\label{coKen}
\mathbf{K} = \frac{1}{2}A^*(S \otimes S^*) A^H, \quad A = \left[\begin{array}{cccc}
1 & 0 & 0 & 1\\
1 & 0 & 0 & -1\\
0 & 1 & 1 & 0\\
0 & j & -j & 0
\end{array}\right]
\end{equation}
where $\otimes$ is the Kronecker product, and  $j = \sqrt{-1}$. Alternatively the Kennaugh matrix for the incoherent case can be obtained from the coherency matrix $\mathbf{T}$ as follows~\cite{You2017}:
\begin{equation}\label{incoKen}
\resizebox{\hsize}{!}{$
\mathbf{K} =\left[\begin{array}{cccc}
\frac{T_{11}+T_{22}+ T_{33}}{2} & \Re(T_{12}) & \Re(T_{13}) & \Im(T_{23})\\
\Re(T_{12}) & \frac{T_{11}+T_{22}-T_{33}}{2} & \Re(T_{23}) & \Im(T_{13})\\
\Re(T_{13}) & \Re(T_{23}) & \frac{T_{11}-T_{22}+T_{33}}{2} & - \Im(T_{12})\\
\Im(T_{23}) & \Im(T_{13}) & - \Im(T_{12}) &\frac{-T_{11}+T_{22}+T_{33}}{2}
\end{array}\right]
$}
\end{equation}
The Kennaugh matrix is real, simple to handle in terms of computation, and it preserves the backscattering information. 

One way of measuring similarity between two Kennaugh matrices utilizes the concept of a geodesic distance. For better contextual understanding the geodesic distance will always refer to the shortest distance on a unit sphere, though it has a much wider connotation~\cite{nakahara2003geometry}. The unit sphere centered at the origin is the locus of points equidistant from the origin with the Euclidean distance equal to unity, i.e. the set $\mathbb{S}^{N-1} = \{(x_1,x_2,\dots x_N) \in   \mathbb{R}^N \mid \sqrt{x_{1}^2 + x_{2}^2 + \dots + x_{N}^2}=1\}$. 
The geodesic distance, denoted as $GD$, between two points $A = (a_1, a_2 \dots, a_N)$ and $B = (b_1, b_2, \dots, b_N)$  on a unit sphere is given by~\cite{Ratha_CD_2017}: 
\begin{equation}\label{GD_th}
GD(A,B) = \cos^{-1}(A \cdot B) = \cos^{-1} \bigg(\sum_{i=1}^{N}a_i b_i\bigg)
\end{equation}

Ratha et al.~\cite{Ratha_CD_2017} discussed the use of geodesic distances between Kennaugh matrices, similarly to~\eqref{GD_th}, by means of
\begin{equation}\label{wf}
GD(\mathbf{K}_1,\mathbf{K}_2) = \frac{2}{\pi} \cos^{-1} \left(\frac{\text{Tr}({\mathbf{K}_1}^T{\mathbf{K}_2})}{\sqrt{\text{Tr}({\mathbf{K}_1}^T{\mathbf{K}_1})}\sqrt{\text{Tr}({\mathbf{K}_2}^T{\mathbf{K}_2})}}\right)
\end{equation}
where $\text{Tr}$ is the trace operator. The factor $2/\pi$ makes the $GD$ for Kennaugh matrices range between $[0,1]$. With this, the distance between Kennaugh matrices is synonymous with the geodesic distance between their projections on the unit sphere in the appropriate dimension ($N = 16$). 

This distance is ideal for characterizing target scattering mechanisms. It is invariant under arbitrary scaling of the Kennaugh matrices, i.e. $GD(\lambda_1\mathbf{K}_1, \lambda_2\mathbf{K}_2) = GD(\mathbf{K}_1,\mathbf{K}_2)$ where $\lambda_1, \lambda_2\in\mathbb{R}$. This property is useful, as the nature of a target does not change under uniform scaling of its Kennaugh matrix. Moreover, $GD(\lambda \mathbf{K}_1,\mathbf{K}_1) = 0$ where $\lambda\in\mathbb{R}$. Thus, $GD$ is positive if and only if the polarimetric nature of the targets is different. 

Similarity and distance (bounded) are complementary quantities. With $GD$ being bounded between $0$ and $1$, $(1 - GD)$ corresponds to a similarity. This implies that the Kennaugh matrix projections away from each other on the unit sphere are dissimilar and those that are nearby are similar. It may be noted that $GD(\mathbf{K},\mathbf{K}) = 0$ implies $\left[1 - GD(\mathbf{K},\mathbf{K})\right] = 1$ which is in line with the definition of similarity. 

This similarity measure is useful for comparing the observed Kennaugh matrix with the one corresponding to a known canonical target. Lee et al.~\cite{Lee2004} used FDD powers to obtain the dominant canonical scattering mechanism from the three-component FDD. The same can be achieved by using the similarity measure proposed in this work. While using a particular decomposition restricts the number of canonical scattering mechanisms considered, the similarity approach allows for any number of desired canonical targets for comparison. Furthermore, on the one hand, the presence of negative power pixels, which is an undesirable phenomenon, occurs in most model based decompositions including the FDD. On the other hand, scattering components using the similarity approach are always non-negative. Hence, using these components instead of the FDD powers as in~\cite{Lee2004}, is a viable option for further study and analysis.

\subsection{Normalized Scattering Similarity Measure}
The odd-bounce scattering which mainly includes the Bragg scattering from the bare ground or the sea surface is modeled using a trihedral corner reflector, and the double-bounce scattering component is modeled by using a dihedral corner reflector. The volume scattering component in this study is modeled as a cloud of uniformly distributed randomly oriented dipole scatterers. This volume scattering model is also used in FDD. The corresponding Kennaugh matrices for the two elementary scatterers, trihedral $(\mathbf{K}_{a})$, dihedral $(\mathbf{K}_{b})$ and the random volume $(\mathbf{K}_{rv})$ are:
\begin{subequations}
\noindent\begin{minipage}{.48\columnwidth}
\begin{equation}
\mathbf{K}_{a}=
\left[\begin{array}{cccc}
1 & 0 & 0 & 0\\
0 & 1 & 0 & 0\\
0 & 0 & 1 & 0\\
0 & 0 & 0 & -1
\end{array}\right]
\label{Eq:trihedral_K}
\end{equation}
\end{minipage}
\begin{minipage}{.48\columnwidth}
\begin{equation}
\mathbf{K}_{b}=
\left[\begin{array}{cccc}
1 & 0 & 0 & 0\\
0 & 1 & 0 & 0\\
0 & 0 & -1 & 0\\
0 & 0 & 0 & 1
\end{array}\right]
\label{Eq:dihedral_K}
\end{equation}
\end{minipage}
\begin{minipage}{1\columnwidth}
\begin{equation}
\mathbf{K}_{rv}=
\left[\begin{array}{cccc}
1 & 0 & 0 & 0\\
0 & 1/2 & 0 & 0\\
0 & 0 & 1/2 & 0\\
0 & 0 & 0 & 0
\end{array}\right]
\label{Eq:random_volume}
\end{equation}
\vspace{0.2mm}
\end{minipage}
\end{subequations}

The similarity measure between an elementary target and the observed Kennaugh matrix is computed from the geodesic distance as,\\
\begin{subequations}
\noindent\begin{minipage}{.49\columnwidth}
\begin{equation}
f_{i}=\left[1-GD(\mathbf{K},\mathbf{K}_{i})\right]
\label{Eq:f_distance}
\end{equation}
\end{minipage}
\begin{minipage}{.49\columnwidth}
\begin{equation}
0 \le \gamma_{i}=\frac{f_{i}}{\sum_{i}{f_i}} \le 1
\label{Eq:gamma}
\end{equation}
\end{minipage}
\end{subequations}
where $\gamma_{i}$ is the normalized similarity with $i\in \{a, b, rv\}$ corresponding to a particular reference target. 

The normalized similarity measure is then modulated with the $\mathrm{Span}$ to make it comparable with FDD scattering powers. Thus, the input $w_{i}$ corresponding to each target $i$, trihedral, dihedral and random volume, is given as: $w_{i} = 2k_{11}\gamma_{i}$, where the first element of the Kennaugh matrix is $k_{11}=\mathrm{Span}/2$. 
In the following, we compare the results of an unsupervised classification scheme using $w_{i}$, i.e. a target corresponding to trihedral, dihedral and random volume, instead of the FDD three component scattering powers as inputs. 
\subsection{Unsupervised Classification}\label{UC}
The unsupervised classification scheme of Lee et al.~\cite{Lee2004} is followed in this work while replacing the FDD powers with the inputs $w_i$ derived in the previous section. The classification steps are summarized as follows:\\ 
\underline{\textit{Preprocessing}}: The coherency matrix obtained from PolSAR data is compensated for orientation~\cite{Lee2000} and subsequently speckle filtered. The inputs $w_a$, $w_b$ and $w_{rv}$ are computed for each pixel. An individual pixel is categorized as $a$ (trihedral), $b$ (dihedral) and $rv$ (random volume) depending upon its maximum similarity to a target (i.e. $\max_{i \in \{a,b,rv\}}({w_i})$). Each category is then further divided into thirty clusters.\\
\underline{\textit{Cluster Merging}}: This process is based on estimating the inter-cluster Wishart distance~\cite{Lee2004} between averaged covariance matrices corresponding to the clusters. This merging is restricted to clusters within the same category $a$, $b$ or $rv$, thus preserving the dominant scattering mechanism. The size of the classes is not allowed to exceed $N_{\max} = 2N/N_d$, where $N$ is the total number of pixels in each scattering category and $N_d$ is the final number of desired classes. \\
\underline{\textit{Wishart Classification}}: By defining the averaged covariance matrix for each class as its center, the Wishart classification is applied iteratively to all the pixels within each category.\\   
\underline{\textit{Output map}}: The Wishart classifier is applied iteratively for a pre-defined number of iterations for convergence while maintaining homogeneity in the classes. We use the standard color convention used in PolSAR for different kinds of scattering: shades of blue are used for trihedral category classes, while red and green are reserved for classes within the dihedral and random volume categories, respectively. 
\subsection{Mixed Category}
The above classification scheme works best if the categorization at the onset is unambiguous. However, there are instances of mixed pixels for which the values of $w_i$ are nearly equal. In such situation, a mixed category pixel is defined as,
\begin{equation}
\frac{\max\{w_a, w_b, w_{rv}\}}{w_a + w_b + w_{rv}} \leq C
\end{equation}
where the threshold $C$ is set at $0.5$. This definition of mixed pixel is similar to the one described in~\cite{Lee2004} using the FDD scattering powers. The unsupervised classification is performed as described in section~\ref{UC}. However, a mixed pixel is allowed to change its category during the Wishart classification stage.	

\section{Results and discussion}
\begin{figure}[htb]
\centering
\subfloat[Pauli RGB]{\label{PauliRGBsf}\includegraphics[height=0.38\columnwidth]{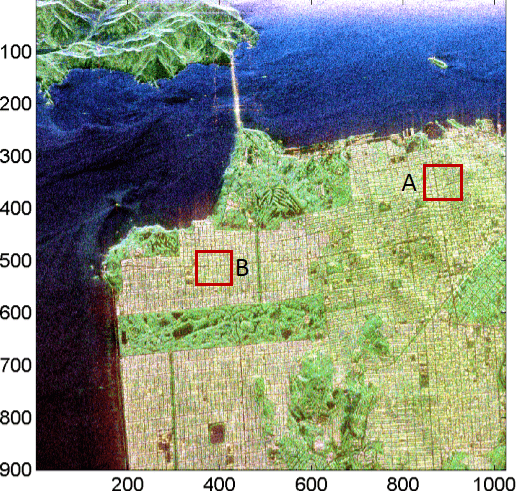}} \hspace{1mm}
\subfloat[Similarity with trihedral $(f_{a})$]{\label{out_ftsf}\includegraphics[height=0.38\columnwidth]{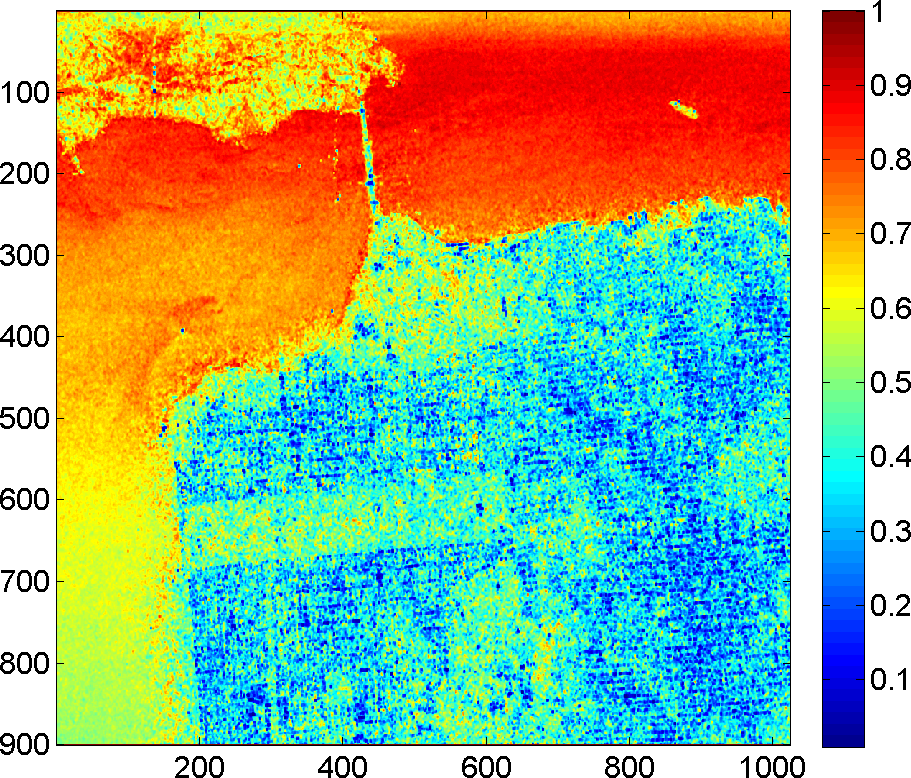}} \\
\subfloat[Similarity with dihedral $(f_{b})$]{\label{out_fdsf}\includegraphics[height=0.38\columnwidth]{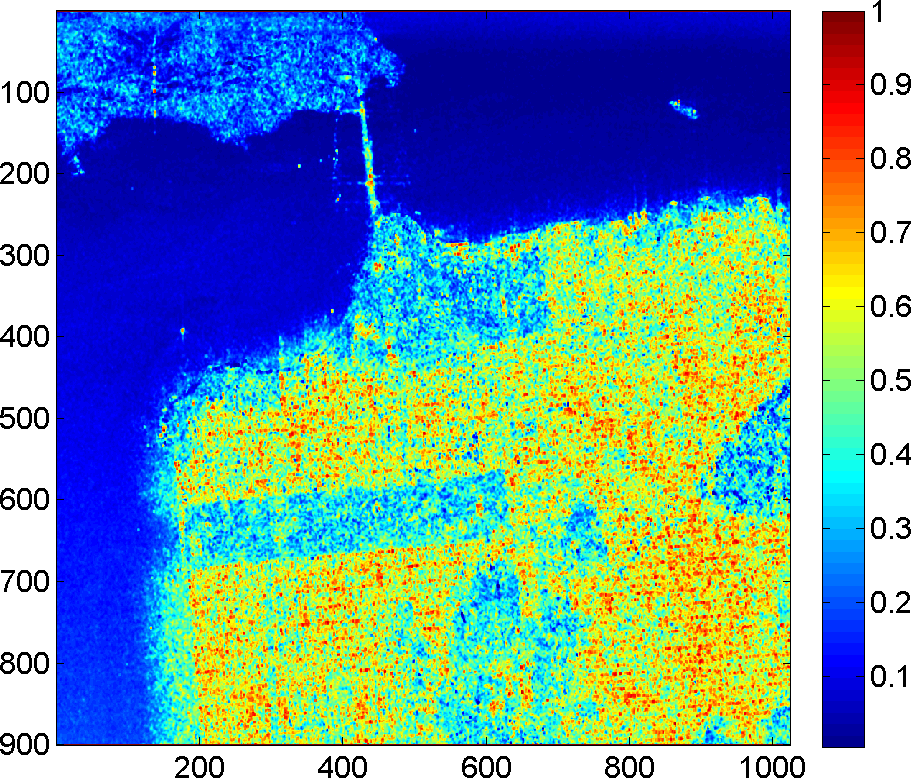}} \hspace{1mm}
\subfloat[Similarity with volume $(f_{rv})$]{\label{out_frvsf}\includegraphics[height=0.38\columnwidth]{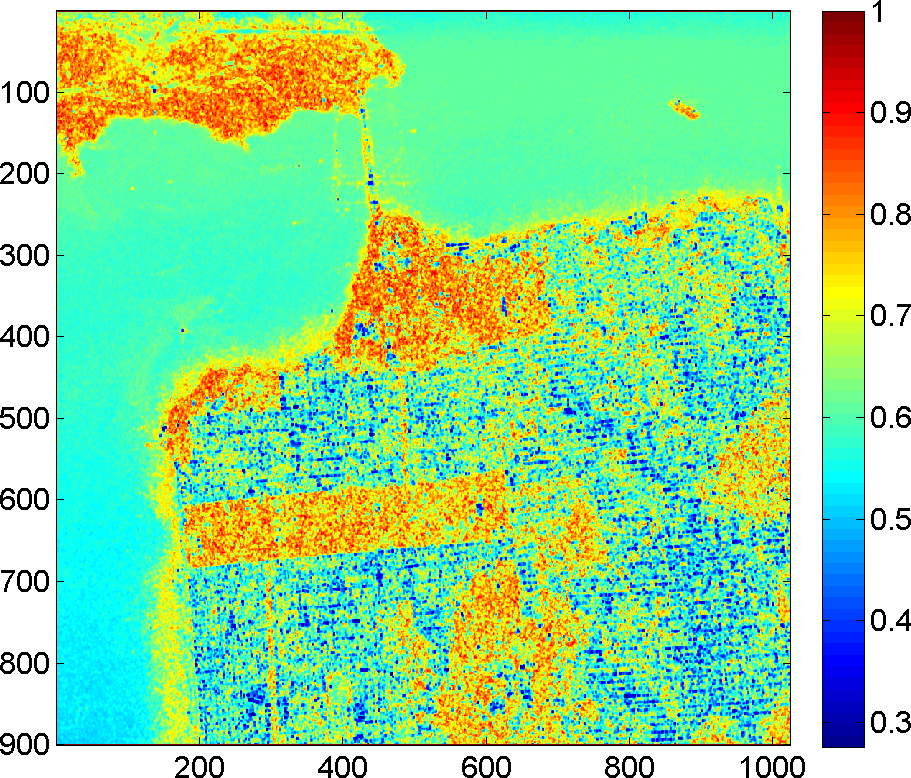}} 
\caption{The Pauli RGB image and the similarity measures $f_{i}$ for the San-Francisco, USA AIRSAR L-band PolSAR data.}
\label{fig:airsar_targets}
\end{figure}

\begin{figure}[!htb]
\centering
\subfloat[Pauli RGB]{\label{PauliRGBmum}\includegraphics[height=0.525\columnwidth]{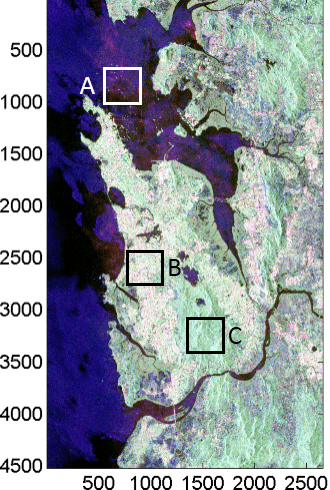}} \hspace{1mm}
\subfloat[Similarity with trihedral $(f_{a})$]{\label{out_ftmum}\includegraphics[height=0.525\columnwidth]{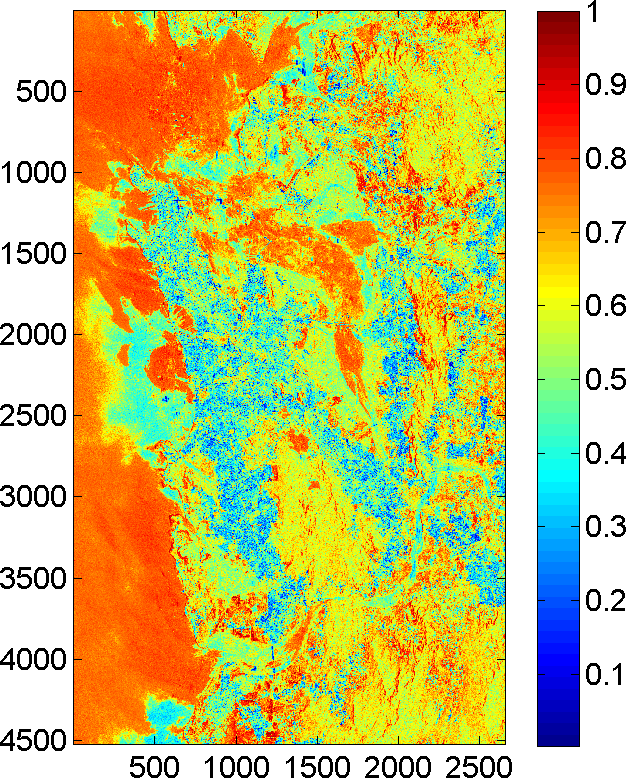}} \\
\subfloat[Similarity with dihedral $(f_{b})$]{\label{out_fdmum}\includegraphics[height=0.525\columnwidth]{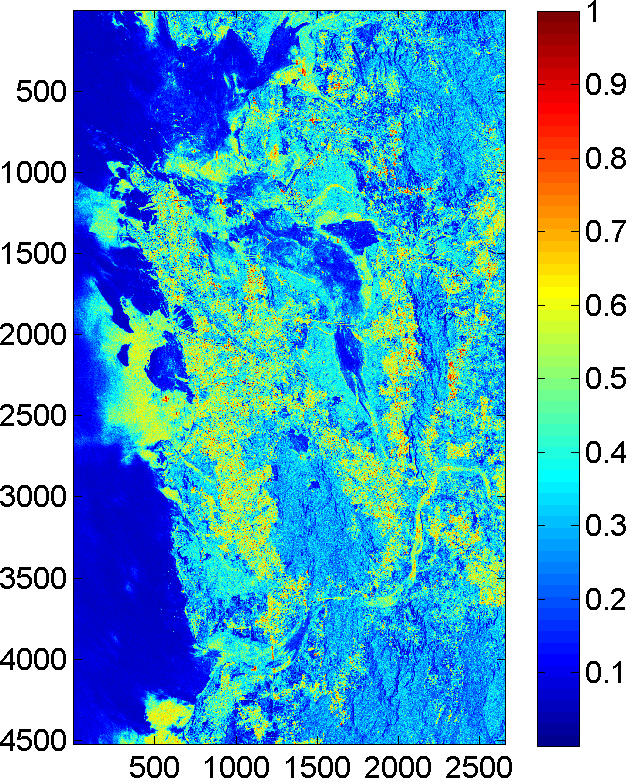}} \hspace{1mm}
\subfloat[Similarity with volume $(f_{rv})$]{\label{out_frvmum}\includegraphics[height=0.525\columnwidth]{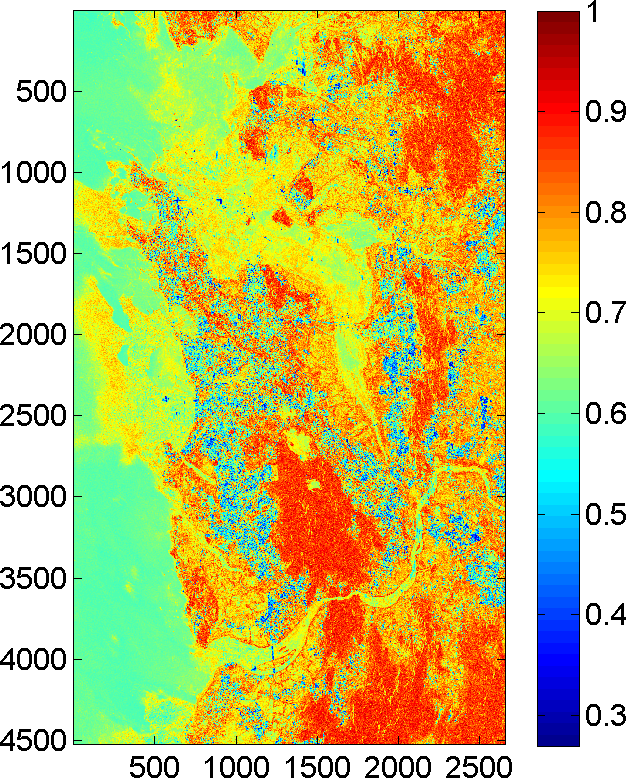}} 
\caption{The Pauli RGB image and the similarity measures $f_{i}$ for the Mumbai, India ALOS-2 L-band PolSAR data.}
\label{fig:alos2_targets}
\end{figure}

We have used a 4-look AIRSAR L-band PolSAR data over San Francisco, the USA, with \SI{12}{\meter} ground resolution with the incidence angle ranging from $5^\circ-60^\circ$. Additionally, an ALOS-2 L-band PolSAR image is used over Mumbai, India with \SI{15}{\meter} of ground resolution. The ALOS-2 image is multi-looked with a factor of $3$ in range and $5$ in azimuth. The Pauli RGB images for the two datasets are shown in Fig.~\ref{PauliRGBsf} and Fig.~\ref{PauliRGBmum} respectively. Certain areas are demarcated in these images for quantitative analysis of classification results. In Fig.~\ref{PauliRGBsf}, A and B are urban areas with different elevation and orientation. In Fig~\ref{PauliRGBmum}, A, B, and C denote ocean, urban and forested areas respectively.

Figs.~\ref{out_ftsf}, \ref{out_fdsf} and~\ref{out_frvsf} show the similarity measures for trihedral $(f_{a})$, dihedral $(f_{b})$ and random volume $(f_{rv})$ for the San-Francisco AIRSAR L-band image, respectively. The similarity measure varies from $0$ to $1$. The $f_{a}$ and $f_{b}$ values are high, in the range of $(0.6,1)$ on water and urban areas respectively. However, to the bottom left corner of the image (on water), $f_{a}$ ranges from $(0.5,0.6)$. 

Interestingly, we observe that $f_{rv}$ values are higher over the water surface $(0.6,0.7)$, even if the $f_{a}$ values are close to $1$. This is due to the fact that $GD(\mathbf{K}_a,\mathbf{K}_{rv})\approx 0.4$ whereas $GD(\mathbf{K}_a,\mathbf{K}_b)=1$. Thus, the Kennaugh matrix structure corresponding to a trihedral target is closer to that of a random volume in comparison to a Kennaugh matrix for a dihedral target under the $GD$ formulation. It can be noticed that the value of $f_{a}$ is high $(0.6,0.9)$ at the top left corner of the image shown in Fig.~\ref{out_ftsf}. The surface scattering is dominant due to high topographic relief with a steep incidence angle of the mountain walls facing the radar.   

\begin{figure}[htb]
\centering
\subfloat[FD-Wishart]{\label{FDWishartsf}\includegraphics[width=0.36\columnwidth]{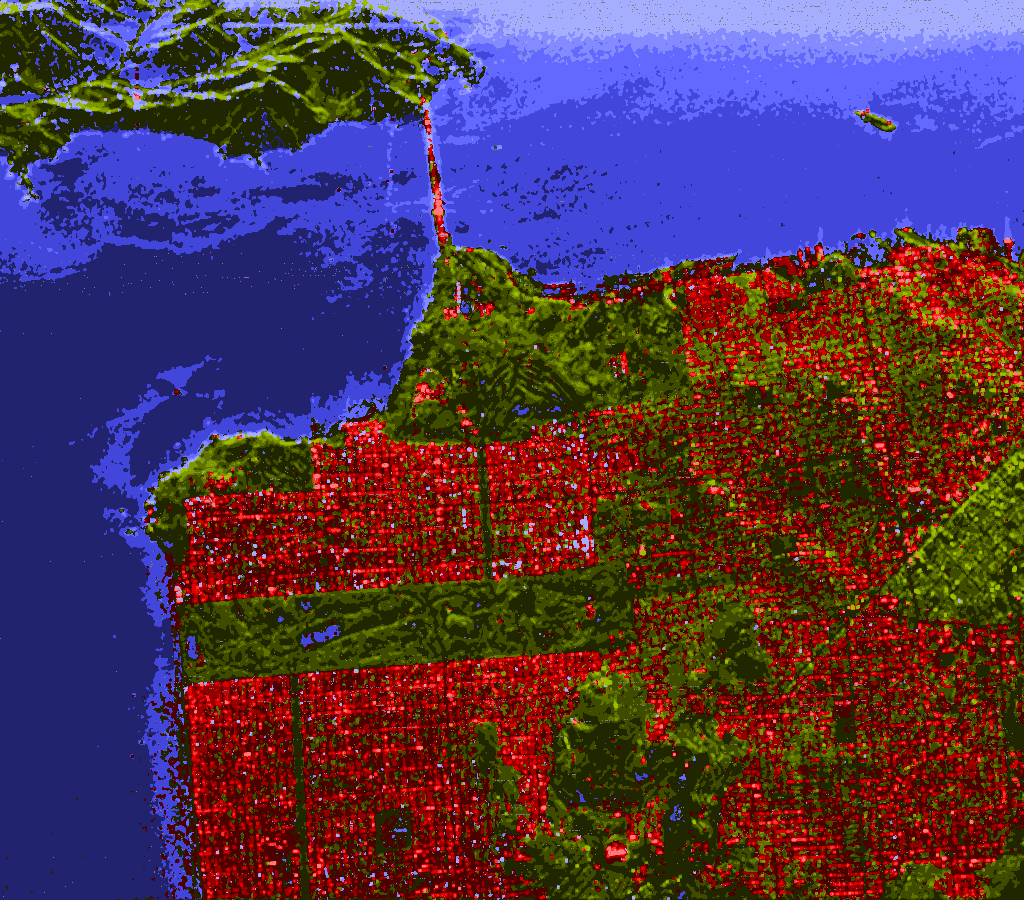}} \hspace{1mm}
\subfloat[GD-Wishart]{\label{GDWishartsf}\includegraphics[width=0.36\columnwidth]{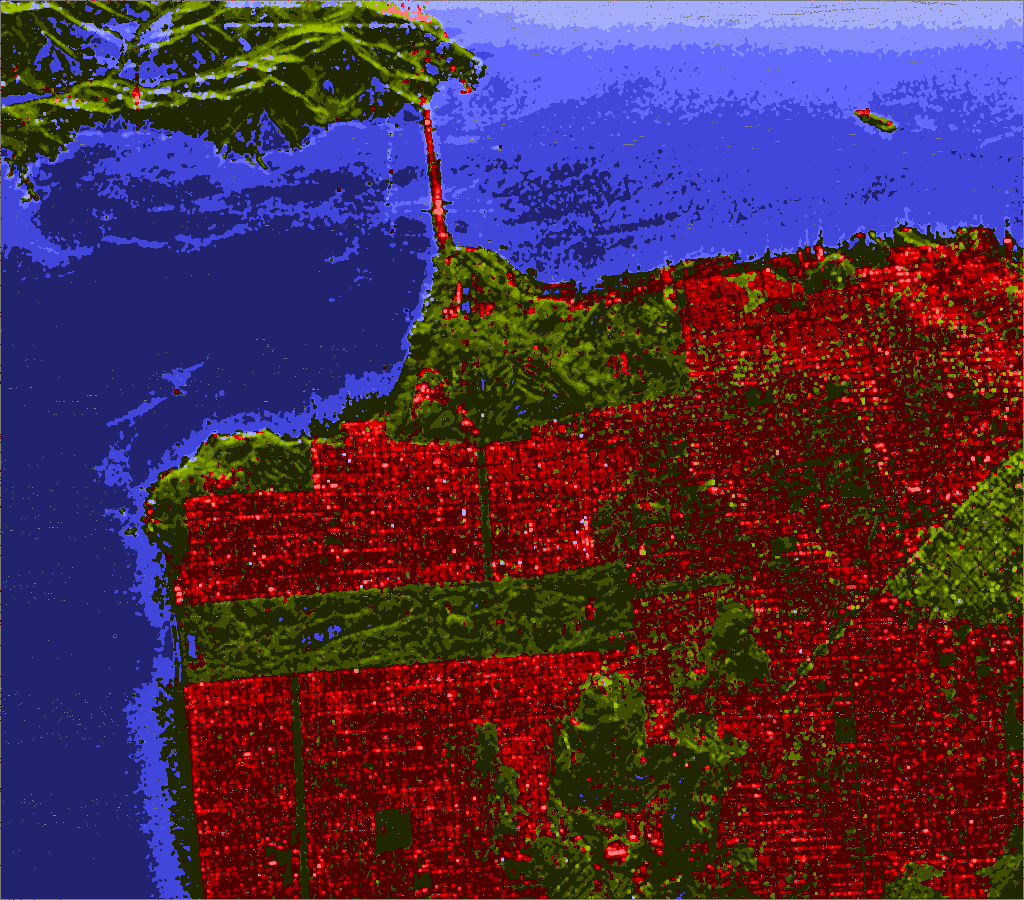}} \\
\includegraphics[width=0.8\columnwidth]{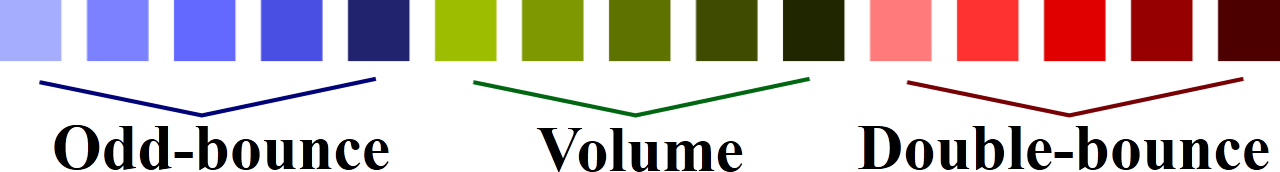} \\
\subfloat[Region A]{\label{comp_sf_a}\includegraphics[width=0.35\columnwidth]{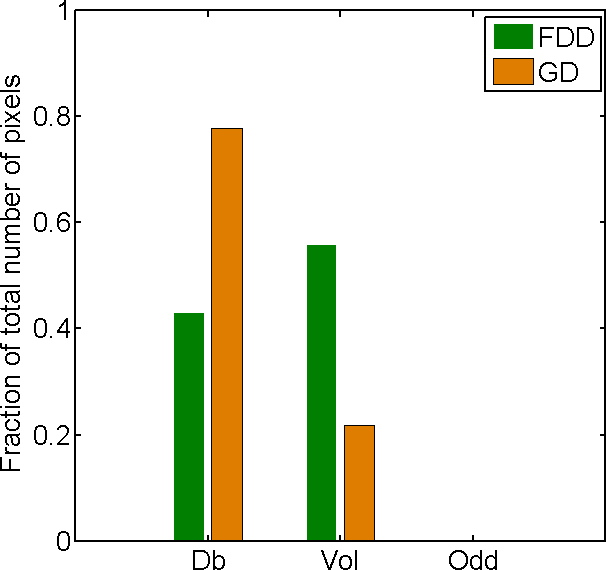}} \hspace{3mm}
\subfloat[Region B]{\label{comp_sf_b}\includegraphics[width=0.35\columnwidth]{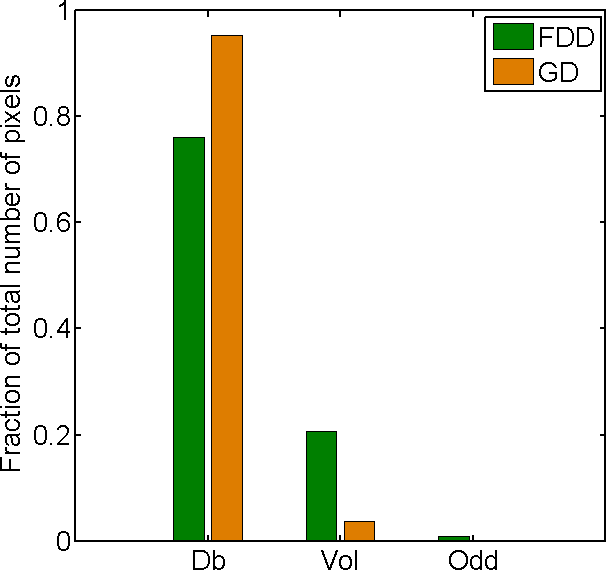}} 
\caption{Classification results using FDD and GD and their comparison for San-Francisco, USA AIRSAR L-band PolSAR data.}
\label{fig:fdd_gd_wishart_sanf}
\end{figure}

Figs.~\ref{FDWishartsf} and~\ref{GDWishartsf} show the results of the unsupervised classification for the San Francisco image using FD-Wishart and GD-Wishart respectively. The image is classified into odd-bounce, double-bounce, and volume scattering categories, each with five classes. It can be seen that the urban area is better classified using the GD-Wishart. Few buildings which are obliquely placed about the radar line of sight are misclassified as volume scatters by FD-Wishart. A quantitative comparison is shown using bar plots in figures~\ref{comp_sf_a} and~\ref{comp_sf_b} for regions A and B. For region A, the double-bounce classification using GD-Wishart is almost double compared to that of FD-Wishart. Similarly, for the region B, the GD-Wishart has the better classification.   

In FDD, the volume scattering power is computed first, only then the residual power from the $\mathrm{Span}$ is redistributed to the canonical scattering mechanisms (odd-bounce and double-bounce). This induces a bias for the volume scattering which could lead to an erroneous classification. Contrarily, the $GD$ weights $(w_{i})$ are computed simultaneously for all the scattering mechanisms (i.e., odd-bounce, double-bounce and random volume). Hence, this restricts the overestimation of the volume scattering component. The classification results are comparable for the both the methodologies on the water surface, forest, and parks.

Figs.~\ref{out_ftmum}, \ref{out_fdmum} and~\ref{out_frvmum} show the similarity measures for trihedral $(f_{a})$, dihedral $(f_{b})$ and random volume $(f_{rv})$, respectively, for the Mumbai ALOS-2 L-band image. On the water class, $f_{a}$ and $f_{b}$ values are complementary, which is in accordance with the orthogonality of the two scattering mechanisms. Such orthogonality translates in the Kennaugh matrix space into maximum possible distance between the canonical scatterers 
(i.e., $GD(\mathbf{K}_a,\mathbf{K}_b)=1$). 

\begin{figure}[!htb]
	\centering
	\subfloat[FD-Wishart]{\label{FDWishartmum}\includegraphics[width=0.26\columnwidth]{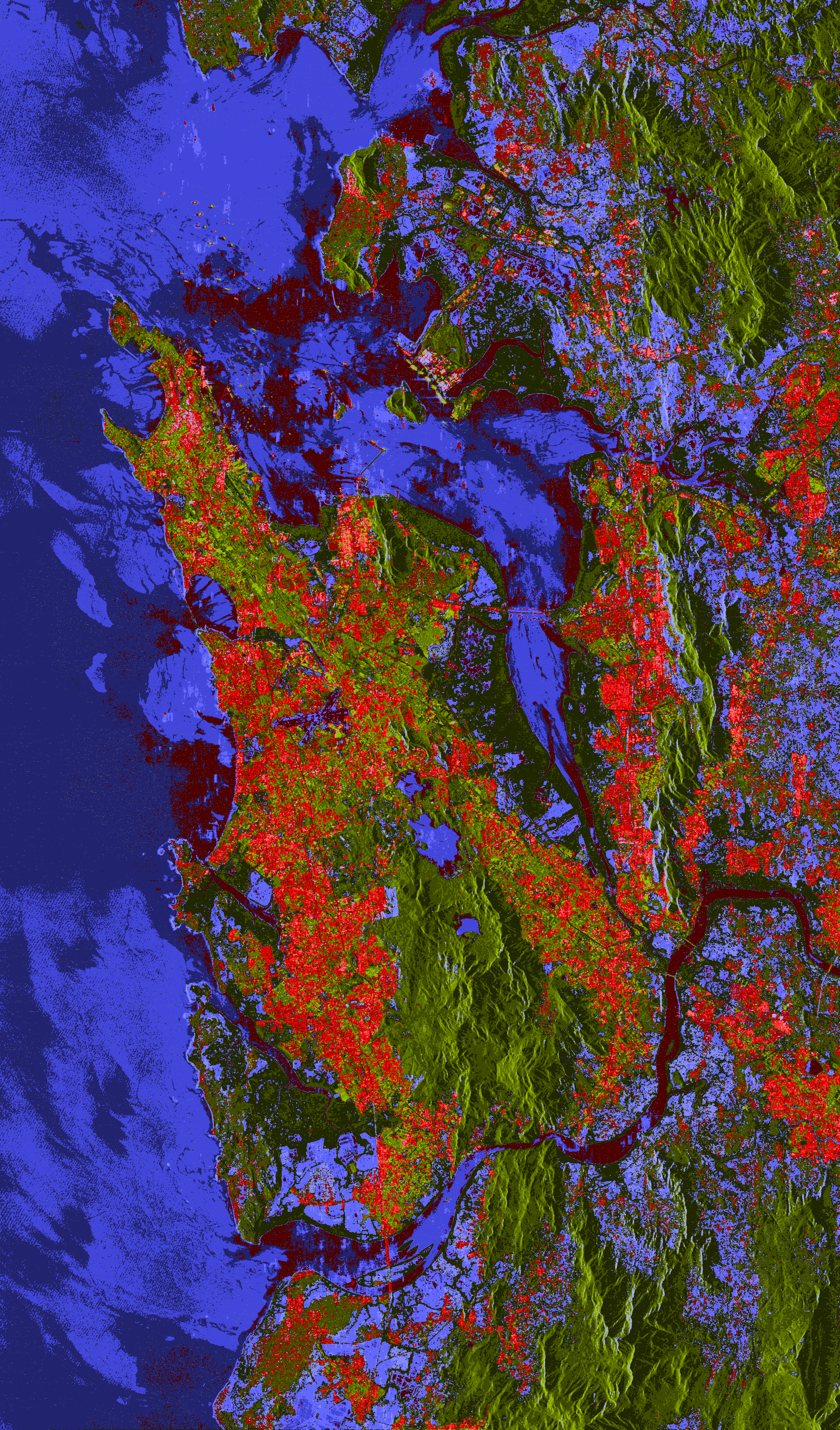}} \hspace{1mm}
	\subfloat[GD-Wishart]{\label{GDWishartmum}\includegraphics[width=0.26\columnwidth]{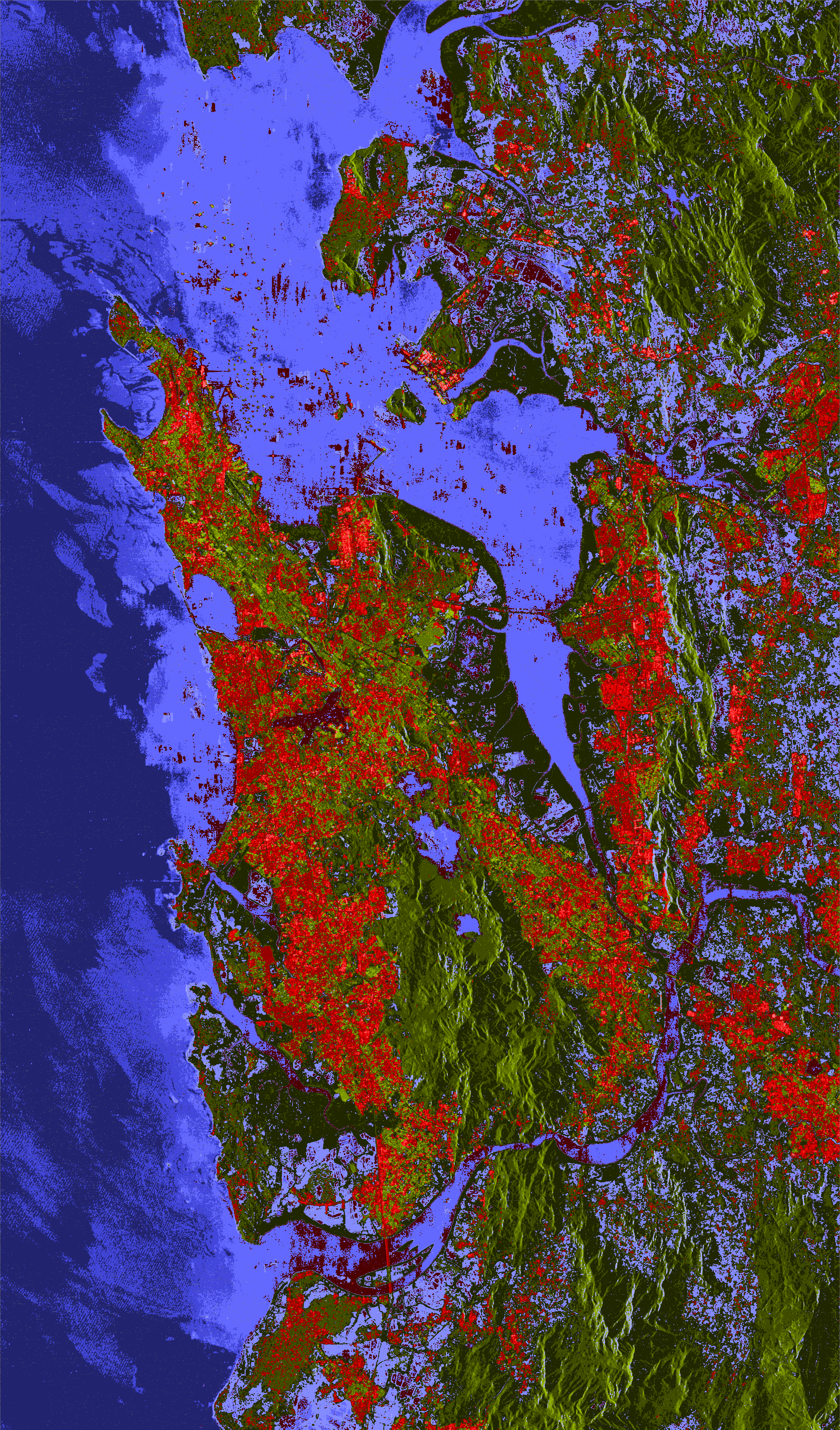}} \\
	\includegraphics[width=0.8\columnwidth]{Legend} \\
	\subfloat[Region A]{\label{comp_mum_a}\includegraphics[width=0.35\columnwidth]{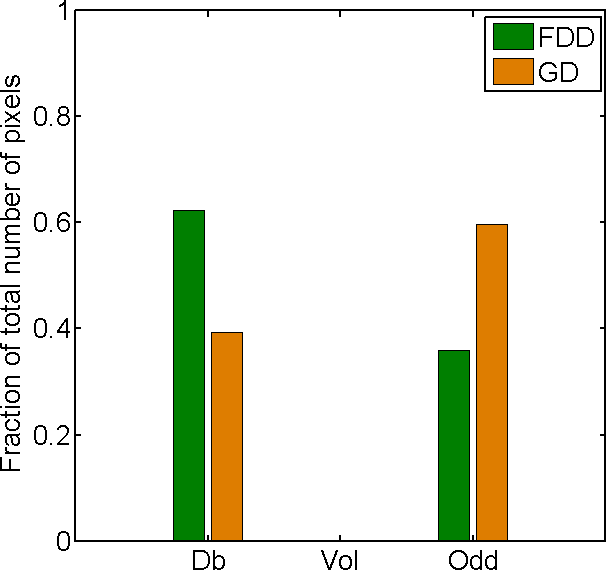}} \hspace{3mm}
	\subfloat[Region B]{\label{comp_mum_b}\includegraphics[width=0.35\columnwidth]{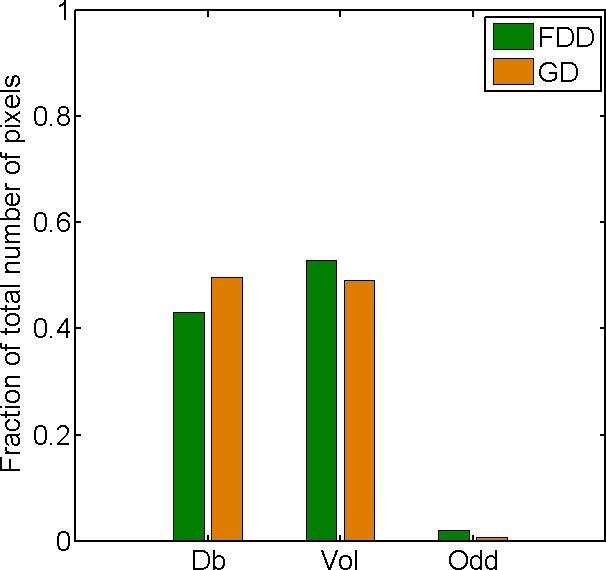}}
	\caption{Classification results using FDD and GD and their comparison for Mumbai, India ALOS-2 L-band PolSAR data.}
	\label{fig:fdd_gd_wishart_mum}
\end{figure}

In Fig.~\ref{out_frvmum}, the vegetation areas are clearly segregated from the surrounding urban area. For example, the forest area C is correctly characterized by high values of $(f_{rv})$. Similarly to the San-Francisco image, the $f_{rv}$ values are moderately high ($\approx 0.7$) over the water surface.

Figs.~\ref{FDWishartmum} and~\ref{GDWishartmum} show the results of the unsupervised classification using, respectively, FDD scattering powers and $GD$ weights $(w_i)$. Again, the image is classified into odd-bounce, double-bounce and volume scattering component each with five classes. The classification of urban areas is improved as one compares GD-Wishart with FD-Wishart. In particular, in FD-Wishart some areas over the ocean surface and rivers (e.g. region A) are misclassified as double-bounce. This is absent in the GD-Wishart. Also in GD-Wishart, the ocean area near and away from the shore separate out better within the trihedral class category. The classification of vegetation areas (e.g. region C) is comparable for both FD and GD Wishart. The differences in classification results are more pronounced for the ALOS-2 image than in the AIRSAR image. 

Figures~\ref{comp_mum_a} and~\ref{comp_mum_b} show the quantitative analysis of classification over regions A and B respectively. Over region A, the double-bounce misclassification is decreased by $\sim20\%$ in GD-Wishart compared to FD-Wishart while there is an increase in the odd-bounce scattering class pixels by a similar amount. Region B shows comparable classification results by both the methods.
\section{Conclusion}
It has been observed that the proposed classification scheme performs better than the FD-Wishart for urban areas. Moreover, proper segregation of different scattering regions over the ocean surface is observed with it. The utilization of the Kennaugh matrix in the proposed methodology makes it suitable for both coherent and incoherent PolSAR datasets. The methodology can be easily extended to more canonical targets. The volume scattering model can be modified from random volume to more advanced models available in the PolSAR literature. The $GD$ may be utilized to replace distances used in classification/clustering algorithms. Naranjo-Torres et al.~\cite{torres2017} discuss a methodology for transforming geodesic distances into test statistics with known asymptotic distribution. This opens a promising avenue for the use of such measures.


%





\ifCLASSOPTIONcaptionsoff
\newpage
\fi



%

\bibliographystyle{IEEEtran}
\bibliography{mybibfile}

%
%


%








\end{document}